\documentclass[11pt,a4paper]{article}
\usepackage[hyperref]{conll-2019}
\usepackage{times}
\usepackage{latexsym}
\usepackage{url}
\usepackage{amsmath}
\usepackage{graphicx}
\usepackage{flexisym}
\usepackage{multirow}
\usepackage{subfig}
\usepackage[colorinlistoftodos]{todonotes}
\usepackage{url}
\usepackage{tikz}
\usepackage{float}

\newcommand{\angry}{\includegraphics[scale=0.23]{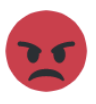}}
\newcommand{\angryy}{\includegraphics[scale=0.23]{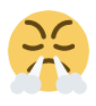}}
\newcommand{\angryyy}{\includegraphics[scale=0.23]{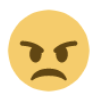}}
\newcommand{\poker}{\includegraphics[scale=0.23]{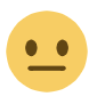}}
\newcommand{\happy}{\includegraphics[scale=0.23]{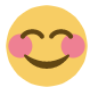}}
\newcommand{\happyy}{\includegraphics[scale=0.23]{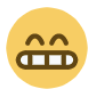}}
\newcommand{\sad}{\includegraphics[scale=0.23]{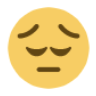}}
\newcommand{\sadd}{\includegraphics[scale=0.23]{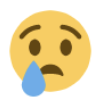}}
\newcommand{\saddd}{\includegraphics[scale=0.23]{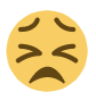}}
\newcommand{\sadddd}{\includegraphics[scale=0.23]{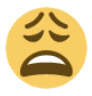}}
\newcommand{\saddddd}{\includegraphics[scale=0.23]{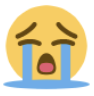}}
\newcommand{\threat}{\includegraphics[scale=0.23]{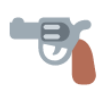}}
\newcommand{\punch}{\includegraphics[scale=0.23]{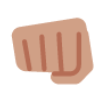}}
\newcommand{\love}{\includegraphics[scale=0.23]{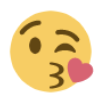}}
\newcommand{\lovee}{\includegraphics[scale=0.23]{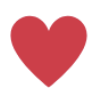}}
\newcommand{\loveee}{\includegraphics[scale=0.23]{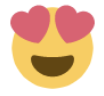}}
\newcommand{\stopp}{\includegraphics[scale=0.23]{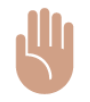}}
\newcommand{\hum}{\includegraphics[scale=0.23]{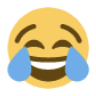}}
\newcommand{\hate}{\includegraphics[scale=0.23]{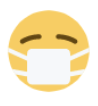}}

\newcommand{\monkey}{\includegraphics[scale=0.22]{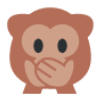}}
\newcommand{\stars}{\includegraphics[scale=0.22]{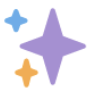}}
\newcommand{\satan}{\includegraphics[scale=0.22]{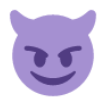}}
\newcommand{\play}{\includegraphics[scale=0.22]{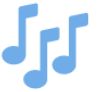}}
\newcommand{\glass}{\includegraphics[scale=0.22]{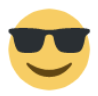}}
\newcommand{\broken}{\includegraphics[scale=0.22]{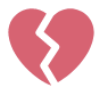}}
\newcommand{\please}{\includegraphics[scale=0.22]{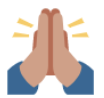}}
\newcommand{\thatsit}{\includegraphics[scale=0.22]{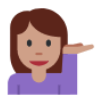}}
\newcommand{\neut}{\includegraphics[scale=0.22]{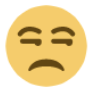}}
\newcommand{\neutt}{\includegraphics[scale=0.22]{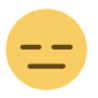}}
\newcommand{\omg}{\includegraphics[scale=0.22]{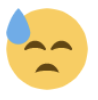}}
\newcommand{\sadddddd}{\includegraphics[scale=0.22]{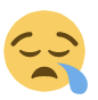}}
\newcommand{\huh}{\includegraphics[scale=0.22]{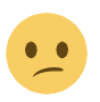}}
\newcommand{\mmm}{\includegraphics[scale=0.22]{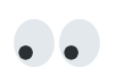}}
\newcommand{\omgg}{\includegraphics[scale=0.22]{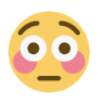}}
\newcommand{\ok}{\includegraphics[scale=0.22]{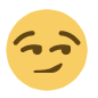}}

\newcommand{\deact}{\includegraphics[scale=0.50]{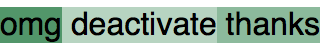}}
\newcommand{\fuck}{\includegraphics[scale=0.50]{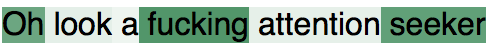}}
\newcommand{\fuckk}{\includegraphics[scale=0.49]{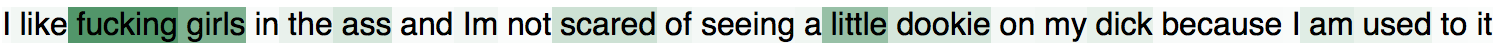}}
\newcommand{\fuckkk}{\includegraphics[scale=0.50]{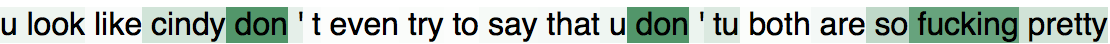}}
\newcommand{\die}{\includegraphics[scale=0.50]{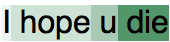}}
\newcommand{\diee}{\includegraphics[scale=0.50]{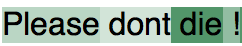}}
\newcommand{\ugly}{\includegraphics[scale=0.50]{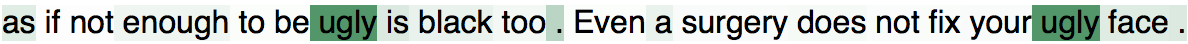}}
\newcommand{\uglyy}{\includegraphics[scale=0.50]{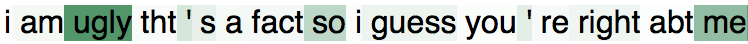}}

\newcommand{\anoy}{\includegraphics[scale=0.50]{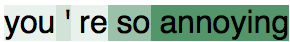}}
\newcommand{\anoyy}{\includegraphics[scale=0.50]{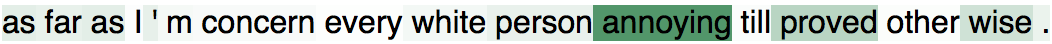}}
\newcommand{\slut}{\includegraphics[scale=0.50]{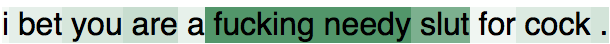}}
\newcommand{\slutt}{\includegraphics[scale=0.50]{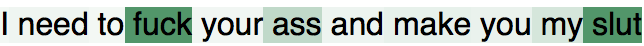}}

\aclfinalcopy 

\title{Attending the Emotions to Detect Online Abusive Language}

\author{Niloofar Safi Samghabadi, Afsheen Hatami, Mahsa Shafaei, \\
\textbf{Sudipta Kar, Thamar Solorio}\\
Department of Computer Science, University of Houston \\
  {\tt \{nsafisamghabadi, amhatami, mshafaei, skar3, tsolorio\}@uh.edu}}

\date{}

\begin{document}
\maketitle
\begin{abstract}
  In recent years, abusive behavior has become a serious issue in online social networks. In this paper, we present a new corpus from a semi-anonymous social media platform, which contains the instances of offensive and neutral classes. We introduce a single deep neural architecture that considers both local and sequential information from the text in order to detect abusive language. Along with this model, we introduce a new attention mechanism called emotion-aware attention. This mechanism utilizes the emotions behind the text to find the most important words within that text. We experiment with this model on our dataset and later present the analysis. Additionally, we evaluate our proposed method on different corpora and show new state-of-the-art results with respect to offensive language detection.
\end{abstract}

\section{Introduction}


Nowadays, abusive behavior has become a serious issue in online societies~\cite{jones2013online, ybarra2004youth}. Unfortunately, such behavior can have serious effects on the physical, mental and social health of the younger generations\footnote{\url{https://enough.org/stats_cyberbullying}}. In recent years, there have been several efforts to automate the detection of offensive language across social media platforms. Traditional approaches using lexical features have been proven to work quite well for this task~\cite{dinakar2012common, davidson2017automated}; however, these types of features can introduce some bias into the system by focusing on profane words, whereas the reports show that most of profanities are used in a neutral way in today's teen talks~\cite{samghabadi2017detecting}. The following examples indicate that profane words are not a good criterion for filtering abusive content anymore:

\noindent\textbf{Neutral: }\textit{Damn you are such a BEAUTIFUL F*CKING MOMMY!}\\
\noindent\textbf{Offensive: }\textit{stop sending questions to yourself pretending you're someone else you weirdo!}

Most of the resources available for this task have been created based on either bad words, or seed words related to abusive topics and do not cover implicit forms of abusive language. In this paper, we propose a method to create a new corpus without focusing on the bad words, and still have a reasonable number of offensive instances. We collect our data from Curious Cat~\footnote{\url{https://curiouscat.me}}, a semi-anonymous question-answering website, that has increased in popularity among teenagers and youth. The anonymity option available on Curious Cat opens the door for digital abuse. On this website, users can choose not to reveal any personal information on their account, as well as post comments/questions on other users' timelines anonymously. 
Due to these properties, we are limited on both the content of a post (since the posts are almost too short in length), as well as the information about the sender of that post.

To overcome the aforementioned challenges within the data, we propose a single deep neural architecture that only employs the textual cues from the input text to decide whether it is offensive or not. Along with this model, we introduce the Emotion-Aware Attention (EA) mechanism that dynamically learns to weigh the words based on the emotions behind the text. We use this method to reduce the model bias towards the bad words. Our main contributions in this paper are as follows:
\begin{itemize}
    \item We create a new corpus for the task of offensive language detection, which is not biased towards profane words.
    \item We propose a neural architecture that captures both local and sequential information from the text to predict whether it is offensive or not. Along with this model, we introduce a new attention mechanism that incorporates emotional information from the text for computing attention weights to find the most important words in the text for the task of offensive language identification. We show that our stacked CNN-BiLSTM model with EA outperforms several strong baselines and also the state-of-the-art across multiple corpora.
    \item We show the effectiveness of our proposed attention model over the regular attention by visualizing the attention weights. We also do an analysis over the mistakes of the model to find if there is any room for improvement.
\end{itemize}

\section{Related Work}
 Abusive language identification and hate speech detection have been addressed by many research papers. Most of the related work have employed feature engineering approaches, and use a combination of different types of lexical, syntactic, semantic, sentiment and lexicon-based features along with classic machine learning algorithms such as Support Vector Machines (SVM), and Logistic Regression~\cite{schmidt2017survey, gitari2015lexicon, van2015detection, davidson2017automated, nobata2016abusive, wiegand2018inducing}.
 
 Due to the popularity of deep neural networks, multiple studies have recently been conducted in order to explore the performance of these models on the task of aggression identification. Most of these studies are focused on hate speech detection within Twitter. ~\newcite{gamback2017using} use a Convolutional Neural Network (CNN) based model, and investigate different textual and embedding features as the input to the model where word2vec produces the best results. ~\newcite{badjatiya2017deep} conduct an extensive evaluation on multiple traditional and deep learning approaches, and report the best results using an ensemble of LSTM and Gradient Boosted Decision Trees.~\newcite{mishra2019abusive} model the network of users with a Graph Convolutional Network (GCN) to learn the structure of online communities along with the linguistic behaviors of the users within them. They report the state-of-the-art results on~\newcite{waseem2016hateful} Twitter data by passing the produced hidden representation by GCN to a Logistic Regression classifier. 
 
 Our model has two major differences with other existing methods: (1) We do not have access to the user-level information due to the nature of the data we work on, and (2) Instead of using an ensemble approach, we propose a single deep neural architecture that shows very promising results across multiple resources. 
 
\section{Dataset}
\subsection{Available Resources}
In this paper, we make use of the following available resources to evaluate our method:
\noindent\textbf{ask.fm}~\cite{samghabadi2017detecting}: Created based on the most frequent profane words in ask.fm and contains around 6K question-answer pairs, where each question and each answer are labeled as positive/neutral or negative.\\
\noindent\textbf{Wikipedia personal attacks dataset}~\cite{wulczyn2017ex}: Includes over 115k labeled discussion comments from the English Wikipedia. This dataset was annotated via Crowdflower annotators, where each label shows if a comment contains a personal attack.\\
\noindent\textbf{Kaggle insult dataset~\footnote{\url{https://www.kaggle.com/c/detecting-insults-in-social-commentary}}}: Released in 2012 for the shared task of ``Detecting Insults in Social Commentary'' hosted by Kaggle, and contains around 6K posts on adult topics like politics, military, etc.

\subsection{New Resource}
We collected our own data from Curious Cat which is a fairly new semi-anonymous, question-answering social media platform, like ask.fm. Curious Cat has been steadily increasing in popularity among teens and pre-teens, and has more than 12 million registered users. This site allows users to post anonymously on the other users' timelines. 

We have crawled around 500K English question-answer pairs from 2K randomly chosen users in Curious Cat. Regarding the annotation process, to avoid having bias in our data, we decided not to focus on profanities. So, instead of using either a dictionary of bad words, or setting seed words related to bullying traces to find the potentially offensive messages, we have chosen to apply a pre-trained classifier with reasonable performance on the other resources. Since the format of the Curious Cat data is similar to ask.fm, we decided to use the state-of-the-art classification method on ask.fm ~\cite{samghabadi2017detecting}. We pre-trained the classifier on the full ask.fm dataset and applied it on Curious Cat in order to automatically label all rows of data. Although ask.fm and Curious Cat have the same format, we notice key differences between these two sources of data, which may affect the quality of automatic labeling substantially. For example, ask.fm data was created based on profane words, so we expected the classifier that was trained on this data to be sensitive to some bad words. However, with Curious Cat, we observed numerous sexual posts that are full of bad words, yet not offensive to the user. In fact, some users encourage others to continue posting sexual comments, like the following example:\\
\noindent\textbf{Question: }\textit{I wanna s*ck your d*ck so hard and taste your c*m.}\\
\noindent\textbf{Answer: }\textit{Enter my DMs beautiful.}\\
Therefore, we created the primary version of our data by randomly selecting 2,482 question-answer pairs, where 60\% were chosen from the negative labeled data, and 40\% chosen from the positive labeled data (we only considered the label of the questions). Then, using a two-way annotation scheme, we asked four annotators to annotate each row of the data to finalize the labels. Table~\ref{data} shows the final distribution of the proposed corpus. The average inter annotator agreement kappa score is 0.499 which shows a moderate agreement among the annotators. It is also interesting to see that 95\% of negative comments were posted on users' timelines anonymously. Table~\ref{compare} compares the four different resources that we use in this paper. We will make our dataset available to the public when the anonymity is not a concern.

\begin{table}[]
\footnotesize
\centering
\begin{tabular}{|l|rrr|}
\hline
Class         & Question & Answer & Total \\ \hline
Offensive     & 609      & 171    & 780   \\ \hline
Neutral & 1873     & 2311   & 4184  \\ \hline
Total         & 2482     & 2482   & 4964  \\ \hline
\end{tabular}
\caption{Curious Cat data distribution~\label{data}}
\end{table}

\section{Methodology}
\begin{figure}[h]
  \centering
    \includegraphics[width=0.48\textwidth, height=0.35\textheight]{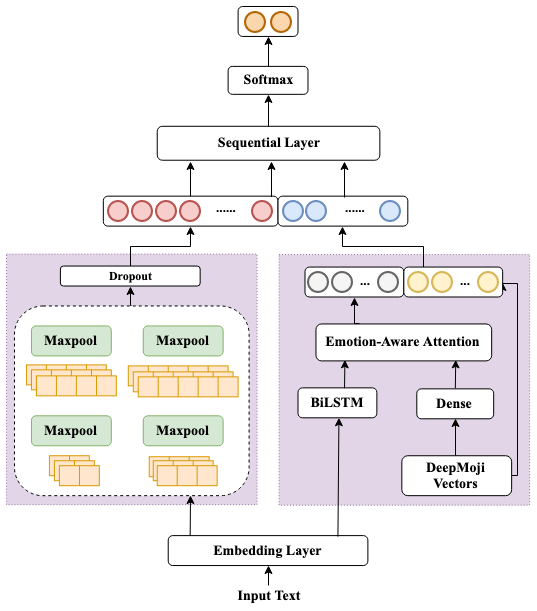}
    \caption{Overall architecture of the model~\label{model}}
    \label{fig:crowd_flower}
\end{figure}

Our proposed model to detect offensive language contains three different modules. The first one uses a Convolutional Neural Network (CNN) to learn the text representation based on local information. CNNs can extract the lexical features, which have been proven to benefit the task of aggression identification. The second module extracts the sequential information from the text via a Bidirectional Long Short-Term Memory (BiLSTM), and uses our proposed Emotion-Aware Attention (EA) mechanism to measure the importance of each word based on the emotions that the text conveys. Essentially, this module extracts the contextual information from the text. The last module is a sequential layer that is applied on top of the output representations from the first and the second modules. It aggregates the lexical and contextual information in order to decide whether the input text is offensive or not. Figure~\ref{model} shows the overall architecture of the proposed model.

\begin{table}[]
\footnotesize
\centering
\begin{tabular}{|l|r|r|}
\hline
Data        & Size       & Negativity ratio \\ \hline
Curious Cat & 4964       & 15.71\%          \\ \hline
ask.fm      & 11194      & 18.08\%          \\ \hline
Wikipedia   & $\sim$115K & 11.70\%          \\ \hline
Kaggle      & 6597       & 26.42\%          \\ \hline
\end{tabular}
\caption{Data comparison~\label{compare}}
\end{table}

\subsection{Embedding Layer}
The input layer of our model is an embedding layer, which takes a sequence of words, extracts the embedding vector for each word, and generates the corresponding embedding matrix for the given input text. We use 200-dimensional Glove~\footnote{\url{https://nlp.stanford.edu/projects/glove/}} embeddings pre-trained on Twitter\footnote{For ask.fm data, we use 300-dimensional Common Crawl Glove pre-trained embeddings, since it works better than the Twitter embedding.}. We also experimented with ELMO~\cite{peters2018deep} and BERT~\cite{devlin2018bert} contextualized embeddings, but the results were not as competative as Glove embeddings. 

\subsection{Convolutional Neural Network (CNN)} 
Several studies on abusive language identification and hate speech detection show promising results with the use of word-level CNN models~\cite{aroyehun2018aggression,zhang2018detecting,zhang2018hate}. Our CNN model takes a sequence of word embedding vectors as input. It contains four stacked 1-dimensional convolution units with 100 filters and different filter sizes of 2, 3, 4 and 5 to extract word n-gram features~\cite{kim2014convolutional}. For each convolution unit, we use ReLU as the activation function, and apply a max pooling operation to take the maximum value as the output feature. Then, we concatenate the outputs for all convolution units and feed the resulting representation to a dropout layer to avoid overfitting. 

\subsection{Emotion-Aware Attention Model (EA)}~\label{emoji}
In addition to the CNN model, we also pass the input embedding vectors to a Bidirectional LSTM (BiLSTM) layer to model the sequential information of the text. Another input to this module is a single emotion vector that is extracted from the input text. For capturing the emotion from the text, we decide to use the DeepMoji model~\cite{felbo2017using} pre-trained on Twitter data. This model creates a representation that contains 64 frequently used online emojis that shows how relevant each emoji is to a given text. We also experiment with NRC emotion lexicon~\cite{mohammad2013crowdsourcing}, but it does not seem to work well with online short texts because of two major limitations: (1) For short texts, the generated vectors are too sparse since the lexicon does not cover many words, and (2) Several words are assigned to more than one category and it would be confusing in the case of short text, since we do not have that much context to decide which emotion is dominant. Inversely, emojis provide us with fine-grained emotional categories that makes this decision easier. 

To prepare the emoji vectors for our model, given a text, we tokenize it to sentences (if it contains more than one sentence). Then, we extract the DeepMoji vector for each sentence and calculate the average vector per post. Ultimately, we make a binary representation that assigns 1 to the five most probable emojis, and 0 to the others. As another input to our model, we pass this emoji vector through a non-linear layer to project it into the same space as the output from the BiLSTM model. Then, both word and emotion representations for the text are fed to the attention model.

The motivation for the EA mechanism is the model presented in ~\newcite{maharjan2018genre}. In that paper, the authors proposed a genre-aware attention model that uses genre information to find the most appropriate set of features for predicting the likability of a particular book. Similarly, we hypothesize that it is not enough to just focus on the word representations in the attention model; because of two reasons: (1) Many bad words are also used in neutral way to make jokes and provide compliments among friends, and (2) Some texts do not contain any profanity, but are still offensive to the receiver. Both reasons may mislead the model to predict the correct label. Therefore, we design the EA mechanism to work not only based on the word representations, but also the emotions behind the whole text in order to better distinguish the most important words of a document. 

Lets assume that $h_i=[\overrightarrow{h_i};\overleftarrow{h_i}]$ is the concatenation of the forward and backward hidden states of BiLSTM, and $e$ is the emoji vector. To measure the importance of words, we calculate the attention weights $\alpha_i$ as follows:
\begin{equation}\label{eq:att_weight}
\alpha_i = \frac{exp(score(h_i, e))}{\Sigma_{i'}exp(score(h_{i'}, e))}
\end{equation}
where the $score(.)$ function is defined as:
\begin{equation}\label{eq:score}
score(h_i, e) = v^Ttanh(W_ah_i + W_ee+b_a)
\end{equation}
where $W_a$ and $W_e$ are weight matrices, and $b$ and $v$ are the parameters of the model. $W_e$ is shared across the words and adds emotion effects to the attention weights. The output of the attention layer is the weighted sum $r$ calculated as follows:
\begin{equation}\label{eq:att_weight}
r = \sum_i\alpha_ih_i
\end{equation}
Finally, we concatenate the output of the attention model with the input emoji vector to further consider the direct effect of the emotions on the model.

\subsection{Sequential Layer}
We concatenate the document representations produced by the two above mentioned modules. The resulting vector is then fed into a hidden dense layer with 100 neurons. To improve generalization of the model, we use batch normalization and dropout with a rate of 0.5 after the hidden layer. Finally, we use a two neuron output layer along with softmax activation to predict whether the input text is offensive or not.

\section{Experiments and Results}

\subsection{Preprocessing and Experimental Setup}

For the Curious Cat dataset, we stratified split the data into train and test sets with a 70:30 training to test ratio, and use 20\% of the training data as the validation set. 
For the other available corpora, we use the same train, validation, and test folds as used by ~\newcite{samghabadi2017detecting}. As for the preprocessing step, we proceed to lowercase the texts and replace all of the links and user mentions with the words ``url'' and ``@username'' respectively. We also truncate the posts to 200 tokens, and left-pad the shorter sequence with zeros.

We use Binary Cross Entropy to compute the loss between predicted and actual labels, and train the network using Adam optimizer~\cite{kingma2014adam} with a learning rate set to $1e^{-5}$. We train the model over 150 epochs, and report the test results based on the best macro F1 obtained from the validation set.

\subsection{Baseline}

We compare our model against the state-of-the-art, as well as several strong baselines listed bellow:

\noindent\textbf{Emoji Baseline: }We use the emoji vectors as the input to this model and directly pass them to the sequential and output layers.\\
\noindent\textbf{CNN: }We feed the output of the CNN module to the sequential and output layers to predict the labels. We also consider concatenating the emoji vectors with the CNN output.\\
\noindent\textbf{BiLSTM + Regular Attention (RA): }Similar to our main model, we use the same BiLSTM module, but calculate the $score(.)$ function without considering emoji vectors using the following formula: $v^Ttanh(W_ah_i+b_a)$. We then feed the resulting representation to the sequential and output layers.\\
\noindent\textbf{BiLSTM + EA: }We use the exact same BiLSTM module as what we have in our main architecture and pass the results to the sequential and output layers.\\
\noindent\textbf{CNN-BiLSTM + RA: }This model is similar to our proposed model, but utilizes the RA mechanism instead of EA.\\
\noindent\textbf{Sam'17: }This model is the state-of-the-art for the ask.fm corpus and is presented in ~\newcite{samghabadi2017detecting}. It makes use of a combination of several textual features as the input to an SVM classifier.\\
\noindent\textbf{Mishra'18: }This model is presented in~\newcite{mishra2018neural} and reported the state-of-the-art results for the Wikipedia dataset. It learns a context-aware representation for characters by concatenating the one-hot character vectors within a document. The resulting representations are then fed to a BiLSTM module with tanh activation, and passed through the output layer.\\ 

\begin{table*}[h!]
\footnotesize
\centering
\begin{tabular}{ll|rr|rr|rr|rr|}
\cline{3-10}
                                                        &         & \multicolumn{2}{c|}{\textbf{Curious Cat}}   & \multicolumn{2}{c|}{\textbf{ask.fm}} & \multicolumn{2}{c|}{\textbf{Kaggle}} & \multicolumn{2}{c|}{\textbf{Wikipedia}} \\ \hline
\multicolumn{2}{|l|}{\textbf{Model}}                              & F1                         & Macro F1       & F1                & Macro F1         & F1                & Macro F1         & F1                 & Macro F1           \\ \hline
\multicolumn{2}{|l|}{Emoji Baseline}                                  &             59.39          &    75.38       &       47.79      &      68.65       &      58.67        &      71.90       &       54.06        &   73.95            \\ \hline
\multicolumn{1}{|l|}{\multirow{2}{*}{CNN}} & -  &  63.62 &   77.31        &        54.91      &       72.99      &      72.08        &       80.88      &        \textbf{79.06}       &         \textbf{88.27}   \\\cline{2-10}
\multicolumn{1}{|l|}{}                  & + emoji &  69.20 &  81.50          & 58.09             & 74.79            & 73.31             & 81.84            & 78.43              & 87.91              \\\hline
\multicolumn{2}{|l|}{BiLSTM + RA}                                  & 64.21                      & 77.87          & 55.16             & 73.08            & 68.85             & 79.35            & 77.29              & 87.29              \\ \hline
\multicolumn{2}{|l|}{BiLSTM + EA}                                  & 65.22                      & 78.63          & 55.27             & 73.26            & 71.05             & 80.53            & 77.05              & 87.14              \\ \hline
\multicolumn{1}{|l|}{\multirow{2}{*}{CNN-BiLSTM + RA}} & -       & 63.84                      & 78.32          & 54.80             & 72.50            & 74.43             & 82.84            & 78.52              & 87.96              \\ \cline{2-10} 
\multicolumn{1}{|l|}{}                                  & + emoji & 69.41                      & 81.55          & 58.55             & 75.07            & \textbf{75.56}    & \textbf{83.56}   & 78.77              & 88.11              \\ \hline
\multicolumn{1}{|l|}{\multirow{2}{*}{CNN-BiLSTM + EA}} & -       & 67.52                      & 80.10          & 55.56             & 72.89            & 73.98             & 82.15            & 78.48              & 87.92              \\ \cline{2-10} 
\multicolumn{1}{|l|}{}                                  & + emoji & \textbf{70.61}             & \textbf{82.41} & \textbf{59.44}    & \textbf{75.40}   & 74.56             & 82.87            & 78.96     & 88.19     \\ \hline\hline
\multicolumn{2}{|l|}{Sam'17}                                      & 65.54                      & 79.03          & 58.47             & 74.09            & 72.85             & 81.73            & 74.48              & 85.76              \\ \hline
\multicolumn{2}{|l|}{Mishra'18}                                   & -                          & -              & -                 & -                & -                 & -                & 77.67              & 87.44              \\ \hline
\end{tabular}
\caption{Classification results in terms of F1 score for the negative/offensive class and macro F1. The results of our proposed model are significantly better than Sam'17 under the Mcnemar significance test.~\label{results}}
\end{table*}

\subsection{Results}

For the evaluation, we use the F1 score for the negative/offensive class, since this is the class of interest. 
We also report the macro F1 score, which calculates the average performance over both classes. This is to ensure that the model does not sacrifice the positive/neutral class to increase the performance of the negative class. 

Table~\ref{results} shows the classification results for all of the resources, where we obtain the best results with our stacked CNN-BiLSTM model for Curious Cat, ask.fm and Kaggle data. Overall, compared to~\newcite{samghabadi2017detecting}, all of the improvements with our model are statistically significant under the Mcnemar significance test ($p-value < 0.001$) for all of the resources. Our model outperforms the-state-of-art for the ask.fm dataset by about 1.5\% with respect to the macro F1. For the Kaggle corpus, we also compare our results against the winner of the Kaggle competition which has the AUC (Area Under ROC curve) score of 0.842~\footnote{\url{https://www.kaggle.com/c/detecting-insults-in-social-commentary/leaderboard}}. The obtained AUC score for this dataset with our model is 0.913, which shows an improvement of 7\%. Only for the Kaggle dataset, the performance of the EA model is slightly worse than the RA. We believe that this is due to the diverse nature of data that we have in different datasets. We will further dig into this analysis in Section~\ref{err}. For the Wikipedia data, we observe that the CNN module obtains the best results (around 1.5\% better than the state-of-the-art F1 score); however, the difference (0.1\%) between the performance of this model and our proposed stacked CNN-BiLSTM is not significant based on the Mcnemar test.

Based on Table~\ref{results}, the CNN module performs better than the BiLSTM module across all of the resources, which proves that the lexical information plays a significant role with respect to offensive language identification. We could also observe that concatenating emoji vectors with the final hidden representation clearly boosts the performance of the system, which confirms our assumption that using the emotional information behind the text benefits the model.  


\section{Why Do the Emoji Vectors Help?}
\begin{figure}[h!]
  \centering
    \includegraphics[width=0.48\textwidth, height=0.12\textheight]{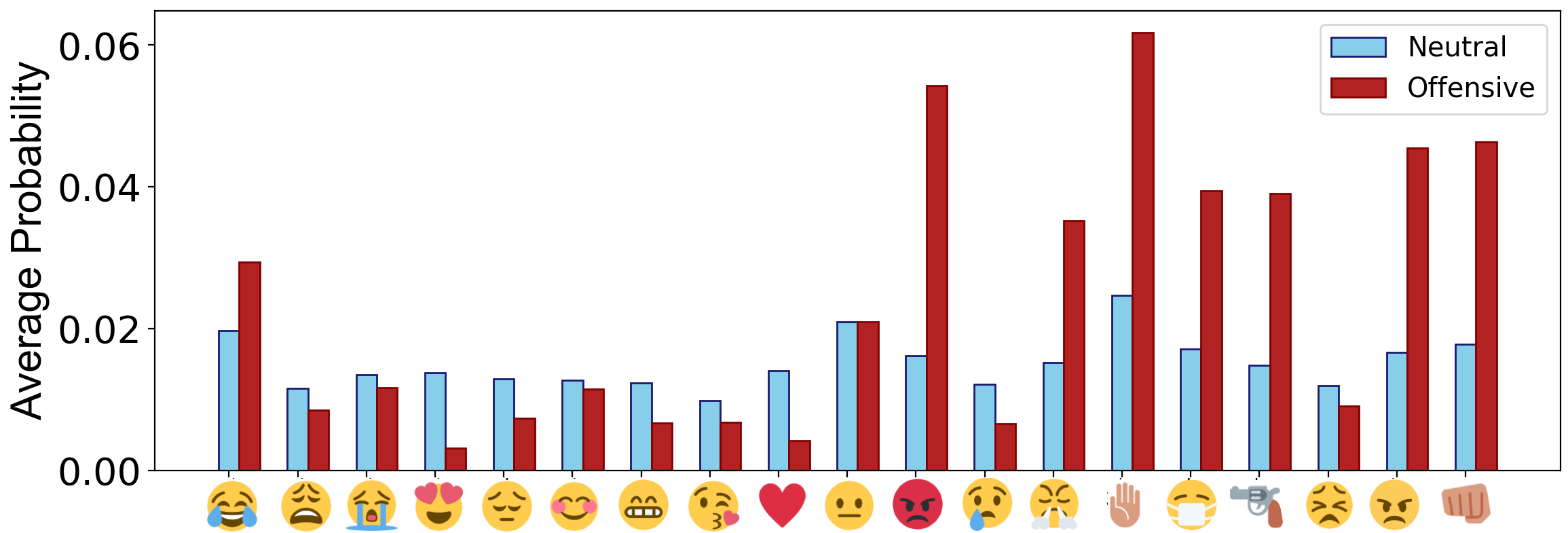}
    \caption{Emoji distribution over Curious Cat data~\label{emoji-dist}}
    \label{fig:crowd_flower}
\end{figure}

Figure~\ref{emoji-dist} shows the emoji distribution over the neutral and offensive classes for the Curious Cat training data. To create this plot, we use the average DeepMoji vector extracted for each instance that shows the relevance of each emoji to a specific comment. 
We create the overall emoji vector per class by averaging the emoji vectors extracted for all of the instances of the same class. Finally, we choose 19 out of the 64 emojis used in the DeepMoji project to create the plot shown in Figure~\ref{emoji-dist}. The fact that there are different patterns visible for the neutral and offensive classes validates our hypothesis on why it is useful to incorporate emoji information to the model. 

\begin{table*}[h!]
\centering
\footnotesize
\resizebox{1\columnwidth*2}{!}{
\begin{tabular}{|l|l|l|}
\hline
 & \textbf{Example} & \textbf{Label} \\ \hline
\rule{0pt}{11pt}\textbf{Ex1}   &\fuck     &  offensive     \\
\textbf{Ex2}   & \fuckkk   & neutral\\
\textbf{Ex3}   & \fuckk   & neutral      \\\hline
\rule{0pt}{11pt}\textbf{Ex4}   & \die  &  offensive     \\ 
\textbf{Ex5}   &  \diee   &  neutral     \\ \hline
\rule{0pt}{11pt}\textbf{Ex6}   & \ugly    &  offensive     \\ 
\textbf{Ex7}   & \uglyy &  neutral      \\ \hline
\rule{0pt}{11pt}\textbf{Ex8}   &  \anoy   &  offensive     \\ 
\textbf{Ex9}   &  \anoyy   &  neutral     \\ \hline
\rule{0pt}{11pt}\textbf{Ex10}   & \slut  &   offensive    \\ 
\textbf{Ex11}   &  \slutt   &  neutral     \\ \hline
\textbf{Ex12} &\deact  &    offensive   \\ \hline
\end{tabular}
}
\caption{Attention visualization for the challenging examples in the Curious Cat data that are correctly classified with our model\label{tab:attn}}
\end{table*}

Based on Figure~\ref{emoji-dist}, angry emojis (\angry, \angryy, \angryyy) are highly correlated with the offensive class, inversely happy and love faces (\happyy, \loveee, \lovee) appeared more frequently in the neutral class. For the happy and love faces \happy and \love, the difference between offensive and neutral classes is much less. We believe that this represents the scenarios where a defender (a user who defends the victim of online attacks) tries to support an attacked user by complimenting him/her, while expressing their hatred towards the attackers. Sad faces (\sadddd, \saddddd, \sad, \sadd, \saddd) are more frequent in neutral instances, which may show the cases where a user expresses his/her unhappiness in response to an attack. It is very interesting that the laughing face, \hum, shows a higher probability for the negative class. This can be linked to the scenario where someone attempts to bully a user by humiliating him/her. Additionally, the plot shows exactly the same probabilities for the poker face (\poker) over the offensive and neutral classes. So, we can conclude that this emoji does not convey any additional information related to offensive language. Other emojis (\stopp, \hate, \punch, and \threat) also seem to frequently appear in the offensive class that indicate the violent and threatening behavior towards the user. 

\section{Attention Visualization~\label{attn-vis}}
Table~\ref{tab:attn} shows the attention visualization for some challenging examples in the Curious Cat test data that are labeled correctly by our best model. We specifically study the instances that are very short in length. The first three rows of the table contain the examples of the word \textit{f*cking} in different contexts. This word is used in two different ways: (1) To express anger, annoyance, contempt, or surprise, or (2) Referring to sexual activities. Looking at Ex1, Ex2 and Ex3, we can see that our model captures these differences. For Ex1, the angry faces are top-rated for the comment. Although, there is no other profane word in the sentence, the model seems to correctly focus on the phrase \textit{attention seeker}. Inversely, for Ex2, \love, \monkey and \stars are listed as the most probable emojis, and the model also weigh the negation word \textit{don't}, and the positive adjective \textit{pretty}. Ex3 shows an instance where the word \textit{f*cking} stands for sexual activities. Top-rated emojis for this comment include \satan, \play and \glass that indicate sexually playful language. We can see that the attention model correctly focuses on the other sexual-related words as well.

Ex4 and Ex5 focus on the word \textit{die}. The first example is obviously offensive towards the receiver. Again, we can see the angry faces plus \threat are extracted as dominant emojis for the text. In this case, \textit{I hope} is also highlighted which seems to trigger the emotions involved in the comment. However, in Ex5, the model also attends to the words \textit{Please} and \textit{dont} which change the emotional direction of the comment to \sadd and \broken and \please emojis. This instance also illustrates that our EA mechanism is able to capture negation in the text. 

The word \textit{ugly} is sometimes used in offensive comments. In our data, we look for examples that only contain this word as the single profane word in order to check whether our model can distinguish between the use of this word in an offensive and neutral way. Ex6 shows a very offensive comment, even though we cannot see any intense bad word in it. It is interesting that our model attends the word \textit{black}, but not the negation words (\textit{not}, \textit{does not}). The top emojis extracted from this text include angry, disgusted and poker faces. Ex7 illustrates the case where the user confirms what the harasser already posted with the hope to prevent further attacks. Unlike the previous example, the dominant emojis for this comment are sad faces along with \broken. It seems that the model captures the self-targeted offensive language by giving a high weight to the word \textit{me} as well as \textit{ugly}.

With respect to the word \textit{annoying}, EA also gives the attention to the words/phrases \textit{you} (Ex8), and \textit{white person} (Ex9), which enables the model to decide if the comment targets a specific person or not. Top rated emojis for Ex8 include the angry faces, but for Ex9 emojis like \thatsit, \neut and \neutt are dominant.

Via Ex10 and Ex11, we could observe that our model is able to distinguish the posts where an intense bad word like \textit{sl*t} is utilized for offending someone (\angry, \angryy, \angryyy) or in a sexually playful conversation (\play, \satan). It is interesting that in Ex11 the weight which is assigned to word \textit{my} is greater than the second person pronouns.

Ex12 displays an instance where there is no bad word in the text, but it is still offensive to the receiver. This example is even more challenging since the comment includes the positive word \textit{thanks}. The attention heat map shows that the top emojis like \stopp, \omg and \sadddddd help the model to learn the negative load of the word \textit{deactivate}.

\section{Error Analysis~\label{err}}


\begin{figure}[h]
\centering
\subfloat[Emoji distribution for correctly classified instances]{%
  \includegraphics[width=0.43\textwidth, height=0.10\textheight]{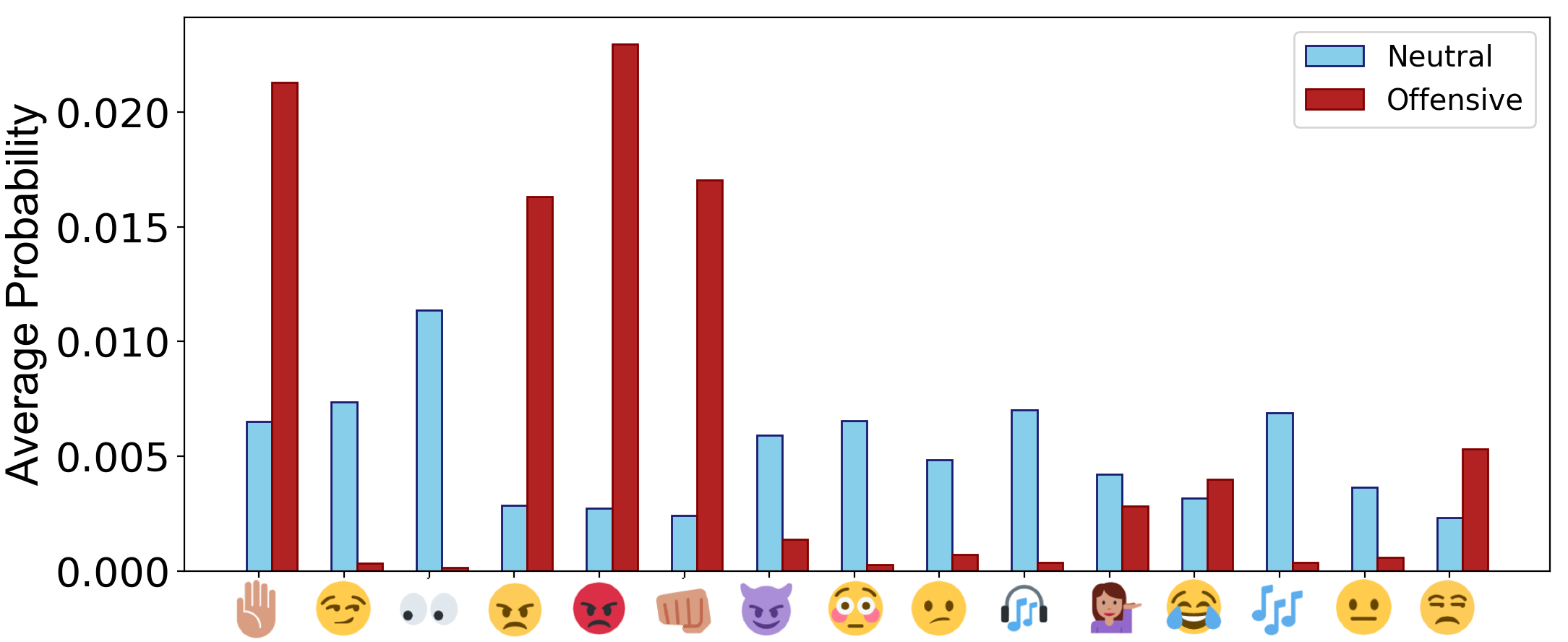}%
  \label{fig:emotionsgroup-a}
}

\subfloat[Emoji distribution for mis-classified instances]{%
  \includegraphics[width=0.43\textwidth, height=0.10\textheight]{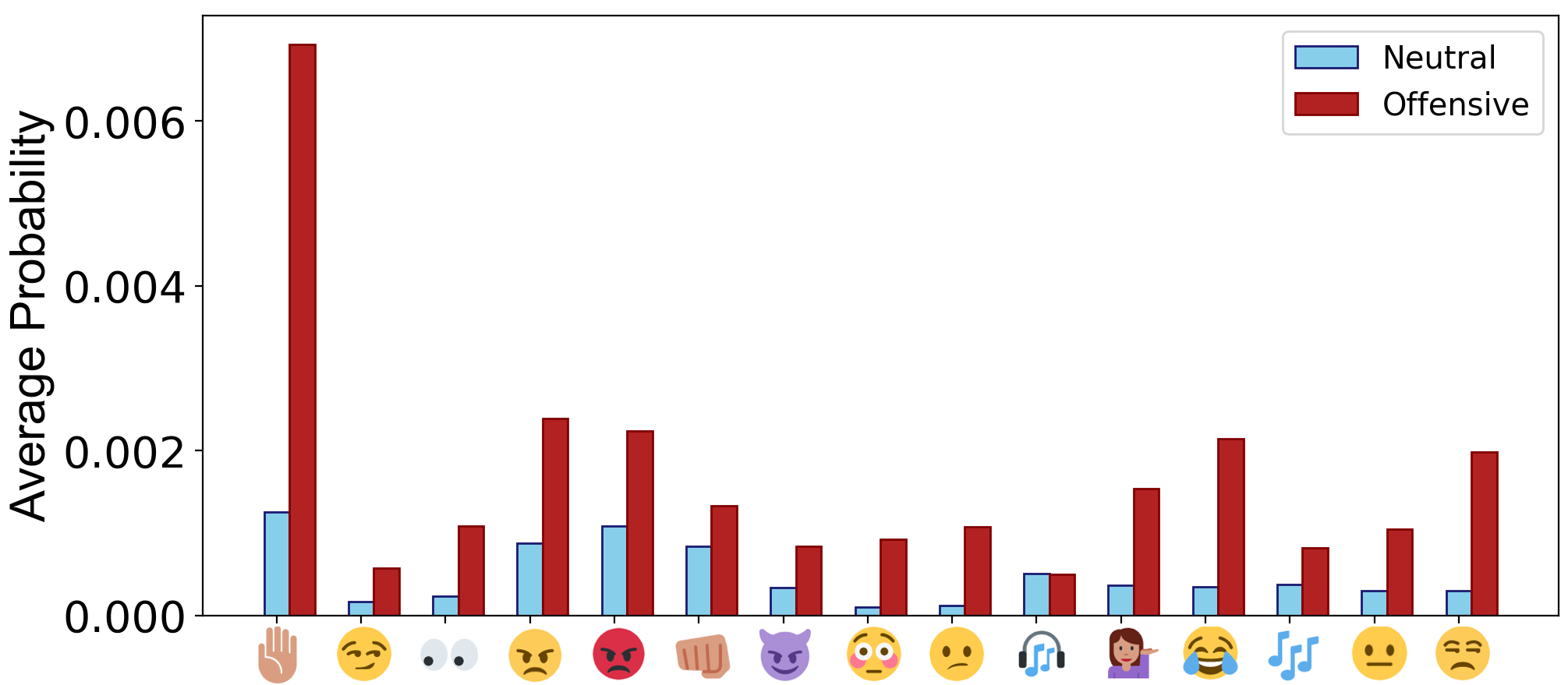}%
  \label{fig:emotionsgroup-b}
}

\caption{Average probability for top emojis extracted from correctly and incorrectly predicted instances of Curious Cat data}
\label{fig:emojigroup}
\end{figure}

Figure~\ref{fig:emojigroup} shows the emoji distribution for the 15 most frequent emojis over the correctly and incorrectly classified instances of Curious Cat test set with our best model. For creating this plot, we calculate the average emoji vector per class for both categories, and we only consider the probabilities for the top five emojis per instance. Based on the two subplots, there is an inverse pattern for some emojis (\ok, \play, \omgg, \huh, \poker, \thatsit, \mmm, and \satan) across the neutral and offensive classes for the correctly and incorrectly classified instances. This may account for most of the classifier's mistakes. We believe that this could be an error which propagated from the DeepMoji model to ours. Besides, comparing the range of the probabilities between the two subplots, the probability scores for the emojis assigned to incorrect predictions are much less than the correct ones. It shows that the DeepMoji model is not confident about the dominant emojis assigned to the mis-classified examples.


Based on the final predictions of our best model on Curious Cat test data, we find that the model is confused when the comment is a question, particularly in cases where the user did not put the question mark at the end of the sentence (e.g. \textit{without sounding like an ignorant dumba*s, what is pansexuality}). We observe several such instances in our data that were labeled as neutral by the annotators, even if there are profane words in the comment. Therefore, it seems that the question statement could change the tone of the language from offensive to neutral. Another source of error are humiliating posts that are very short and do not contain profanity (e.g. \textit{Fix your teeth}). We believe that for detecting this kind of offensive instances, we need to also consider the answer that the user provides for the received comment.

On the other hand, we are also interested to investigate the reasons behind the superiority of RA model to EA when it comes to the kaggle data. Looking at the mislabeled instances by EA, which are labeled correctly with RA, we find that our EA model is not able to correctly label general questions like ``Why you gotta trash Cali?'' (actual label = neutral, predicted label = negative). It is also unable to detect the insults that are indirectly addressed to the user (e.g. \textit{It must suck to be so stupid. mindless , though-controlled libturd sheeple}), and some offensive slang (e.g. \textit{Back under your rock}), since the DeepMoji model assigns the irrelevant emojis to them.

\section{Conclusion and Future Work}
In this paper, we propose a stacked CNN-BiLSTM model with an emotion-aware attention mechanism as a new architecture to detect online abusive behavior and hateful language. We make use of DeepMoji vectors to extract the emotion behind the text, and show it's major effects to benefit the performance of the model through the analysis section. Using our proposed model, we outperform the state-of-the-art results and several strong baselines across the three existing corpora. We also create a new resource for the task of detecting offensive language that does not focus on bad words. Our model shows very promising results over this dataset as well. 

As for the future work, due to the fact that perceived level of aggression is very subjective to the user, we plan to jointly model the question and answer within a pair for the Curious Cat and ask.fm data. We believe that the reply that the user provides in response to a received question/comment is a strong indicator whether it was offensive or neutral towards the user. Another possible path in order to move the research forward, is to expand this task to the detection of cyberbullying episodes which have become a growing concern in online societies.

\bibliography{conll-2019}
\bibliographystyle{acl_natbib}

\end{document}